\def\year{2022}\relax
\documentclass[letterpaper]{article} 
\usepackage{aaai22}  
\usepackage{times}  
\usepackage{helvet}  
\usepackage{courier}  
\usepackage[hyphens]{url}  
\usepackage{graphicx} 
\usepackage{natbib}  
\usepackage{caption} 
\usepackage{subcaption}
\DeclareCaptionStyle{ruled}{labelfont=normalfont,labelsep=colon,strut=off} 
\usepackage{algorithm}
\usepackage[noend]{algpseudocode}
\usepackage{amsmath}
\usepackage{amssymb}
\usepackage{amsfonts}
\usepackage{amsthm}
\usepackage{booktabs}
\usepackage{tikz}
\usepackage{hyperref}
\usetikzlibrary{arrows}
\usepackage{pgfplots}
\usepgfplotslibrary{groupplots}
\usepackage{graphicx}
\usepackage{textcomp}
\usepackage{xcolor}
\theoremstyle{definition}

\pgfdeclareimage[interpolate,height=3mm,width=3mm]{ghost_blue}{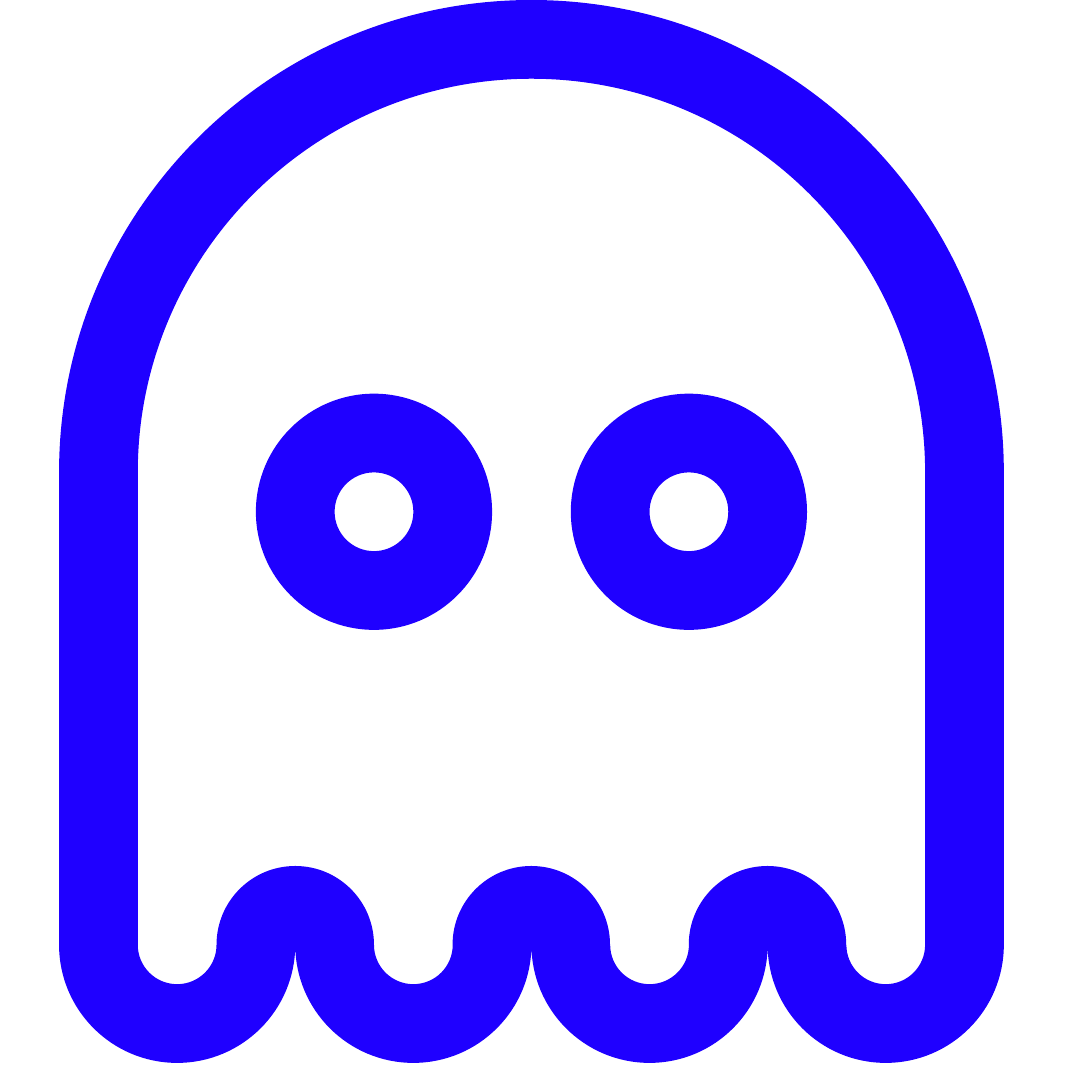}
\pgfdeclareimage[interpolate,height=3mm,width=3mm]{ghost_red}{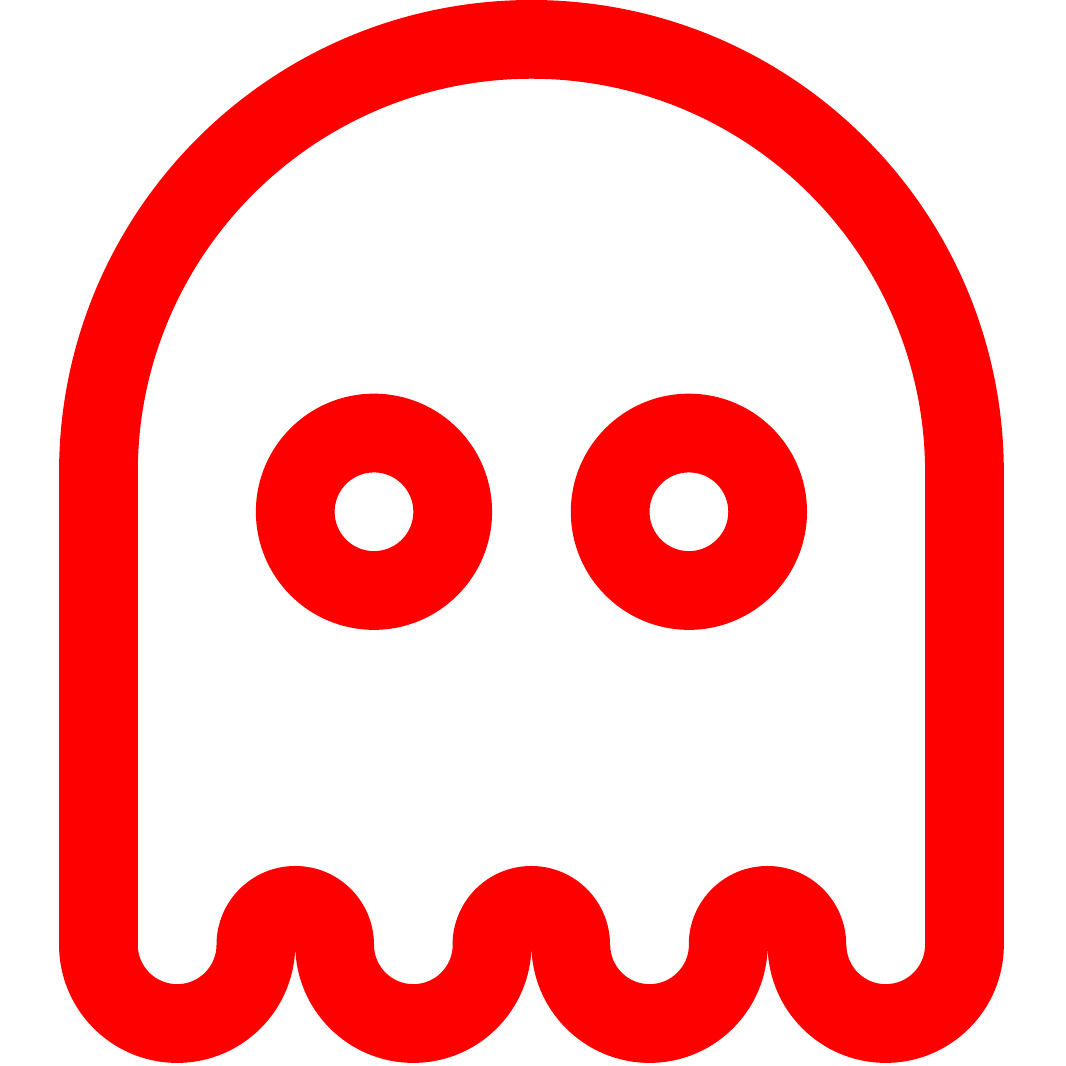}
\pgfdeclareplotmark{ghost_blue}{%
\pgftext[at=\pgfpoint{0pt}{1pt}]{\pgfuseimage{ghost_blue}}%
}
\pgfdeclareplotmark{ghost_red}{%
\pgftext[at=\pgfpoint{0pt}{1pt}]{\pgfuseimage{ghost_red}}%
}
\title{Explaining Deep Learning Representations by Tracing the Training Process}
\author{
    Lukas Pfahler,
    Katharina Morik
}
\affiliations{
	TU Dortmund University,\\
	Dortmund, Germany\\
	lukas.pfahler@tu-dortmund.de
}

\begin{document}

\maketitle

\begin{abstract}
We propose a novel explanation method that explains the decisions of a deep neural network by investigating how the intermediate representations at each layer of the deep network were refined during the training process.
This way we can a) find the most influential training examples during training and b) analyze which classes attributed most to the final representation.
Our method is general: it can be wrapped around any iterative optimization procedure and covers a variety of neural network architectures, including feed-forward networks and convolutional neural networks. We first propose a method for stochastic training with single training instances, but continue to also derive a variant for the common mini-batch training.
In experimental evaluations, we show that our method identifies highly representative training instances that can be used as an explanation. Additionally, we propose a visualization that provides explanations in the form of aggregated statistics over the whole training process.
\end{abstract}
\section{Introduction}
Contemporary approaches for explainability in machine learning focus either on explaining individual predictions or the full model \cite{Langer/etal/2021a,Selbst2018}. 
Explanation methods should answer the question "Why did the model compute this output?". 
Many methods approach this question by providing a description on what the model does or how it computes its outputs. 
Selbst and Baricas observe that
\begin{quote}
''so far, the majority of discourse around understanding machine learning models has seen the proper task as opening the black box and explaining what is inside.`` \cite{Selbst2018}
\end{quote}
For instance for convolutional neural networks, saliency methods \cite{Simonyan13} provide explanations in the form of heatmap overlays for the input images highlighting which region was most important to the convolutional neural network. Then it is left to the user to infer or interpret why that region is important for the decision. Unfortunately, these methods often show little dependence on the model they try to explain \cite{Adebayo/etal/2018a}. 

We argue that, in the case of machine learning models, the leading question for designing explanation methods should be extended to "Why did the model \emph{learn} to compute this output?
The key to answering that question lies in the training process of the model. For instance a decision tree can provide evidence for its decision by reporting the class frequencies that fall into nodes along its decision path.
In this work we aim to provide similar evidence for the decisions of deep networks.





\begin{figure}
\includegraphics[keepaspectratio, width=\linewidth]{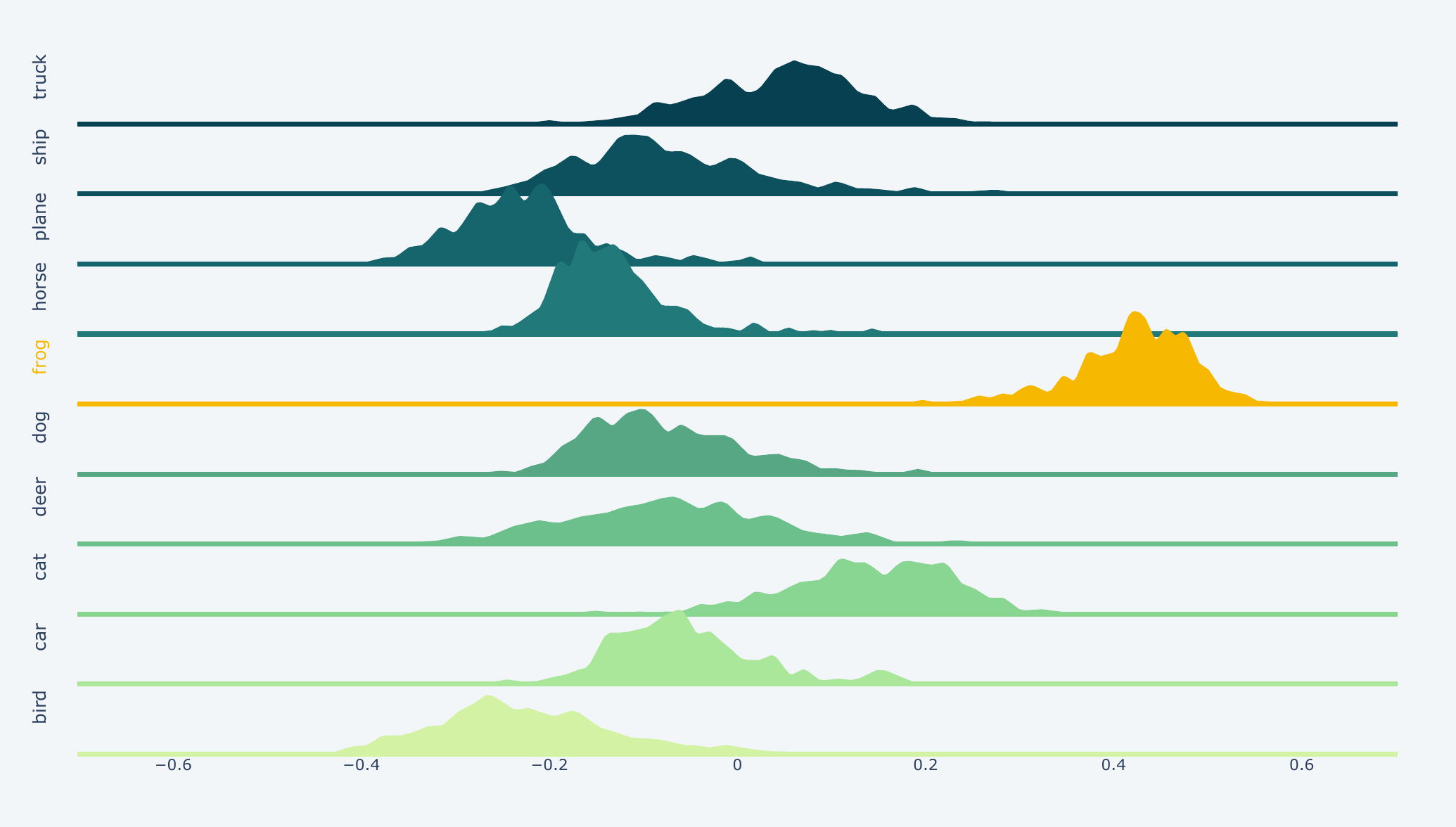}
\caption{Example Explanation: ''\textit{The model predicts \textbf{frog} as training examples of that class had the highest influence on the computed intermediate representation in layer 14}``.}
\label{fig:example}
\end{figure}

We propose a method for tracking the training process of deep networks.
By archiving all weight updates of a deep network, we can attribute the representations computed at each layer to individual training examples.
Ultimately we can support the explanation of a model's decision with data points in the training data.
To this end, we propose two visual explanations: First, we can identify and show the most-influential training examples, second we can visualize how all training examples of a class contributed to the computed representaion in ridge plots
like in Figure~\ref{fig:example}.

The rest of this work is structured as follows: We begin with a discussion of related work on explanations, focussing on methods that incorporate the training data or the training process in its explanations. In the Method section, we present our method for training deep networks that can provide explanations which is based on tracking the training process. We begin by introducing our idea for any stochastic optimization method without mini-batching, but proceed to develop a training approach that works well with mini-batching.
In our experiment section, we evaluate our approach in two classification tasks and investigate the computational overhead. We conclude this work with a brief overview and discussion of future research questions.

\section{Related Work}
There is a rich body of related work on methods for explaining machine learning models \cite{Rudin/etal/2021a,Beckh/etal/2021a}. We focus on works that explain individual decisions rather than the full model.
An important branch of explanation methods focusses on "opening the black-box".
Attribution based methods try to identify the most important input dimensions that lead to a decision. For convolutional neural networks, this includes methods like saliency maps\cite{Simonyan13} and other backpropagation-based approaches\cite{Nie/etal/2018}. 
Another important branch of research focusses on feature importances that estimate which features influenced a decision the most. Prominent approaches include Shapley values (see e.g. \cite{Molnar/2019}) and SHAP\cite{Lundberg/Lee/2017}. 
Our approach, however, is more related to example-based explanation methods.
Using the training data in explanation methods is not novel. In the simplest form, nearest-neighbor based classifiers provide explainable decision through the nearest neighbor examples \cite{Molnar/2019}.
Support vector machines have a similar property: Looking at the support vectors -- all of them or only the most influencial onces for a particular decision -- provides interpretable models or explainable decisions \cite{Rueping/2006}.

Permutation based approaches are an easy way to measure the importance of individual training instances to the final model by dropping them from the training data and monitoring the change of the model. However they do not scale to large-scale deep learning problems.
Influence Functions\cite{Koh/Liang/2017} provide an alternative. They investigate how the model changes when individual examples are reweighted in the empirical loss function. This can be approximated by investigating the Hessian matrix of the loss function around the final model. For interesting models, however, this Hessian matrix can only be approximated\cite{Koh/Liang/2017}. Yeh et al. \shortcite{Yeh/etal/2018a} also propose to decompose pre-activations of neural networks to generate explanations, however they decompose them based on the training data, while we decompose based on the training \emph{process}. 

Perhaps most relevant to this work are methods that also trace the training process of gradient-based methods: Garima et al. \cite{Garima/etal/2020a} decompose the losses of training instances to sums over the losses of all intermediate models during training. As in our approach, this requires storing parameters in all training steps: While we store the parameter changes, the authors store the intermediate models. F
Their method explains the final decision of the network, while we also provide explanations for intermediate representations of a neural network.
Follow-up work by Zhang et al. \cite{Zhang/etal/2020a} extends this work for natural language classifiers which allows finer-grained contribution of the influence to spans of training texts, rather than the full text.
Swayamdipta \cite{Swayamdipta/etal/2020a} analyze the training process, but their goal is to identify easy and hard training instances in the training data to help the machine learning engineer assess the data quality. 

Finally we should note that evaluating explanations is a non-trivial task on its own \cite{Zhou/etal/2021a}.
Particularly visual explanations and example-based explanations \cite{Nguyen/Martinez/2020a} are often subjective and defining quantitative metrics is difficult.

\section{Method}
\label{sec:method}
Stochastic gradient descent (SGD) and its variant are the most popular and most frequently used learning algorithms for a diverse range of models, most notable deep networks, but also for kernel methods \cite{Pfahler/Morik/2018} or matrix factorization methods \cite{Hess/2018}. For supervised learning, SGD trains a model $f_\theta$ parameterized by real-valued weights $\theta$ by minimizing the empirical risk
\begin{equation*}
\arg \min\limits_\theta \frac 1 n \sum\limits_{i=1}^n \ell(f_\theta(x_i), y_i).
\end{equation*}
Optimization is performed in $T$ iterations, where, at the $t$-th iteration, we sample an example $(x_t, y_t)$ from the training data and update the weights in the negative direction of the gradient of the loss 
\begin{equation}
\theta^{(t+1)} = \theta^{(t)} - \eta^{(t)} \nabla_\theta \ell^{(t)}(f_{\theta^{(t)}}(x_t), y_t)	
\label{eq:sgd}
\end{equation}
with a small step-size $\eta^{(t)}$. 
Hence the output of the SGD training algorithm is a vector of weights that is the sum of all weight updates performed during training
\begin{equation}
	\theta^* = \sum\limits_{t=1}^T \underbrace{-\eta^{(t)} \nabla_\theta \ell^{(t)}(f_{\theta^{(t)}}(x_t), y_t)}_{=:\Delta^{(t)}}.
	\label{eq:core}
\end{equation}

For the purposes of this paper, we can view this as a sum of vectors $\Delta^{(t)}$. Hence variants of SGD like SGD with momentum \cite{Sutskever/etal/2013a}, Adam \cite{Kingma/Ba/2015a} or SAM \cite{Foret/etal/2021a} also fit within this framework. These variants do not use the plain gradients, but update the weights based on a function of the gradient and an internal state of the optimizer. In this case, we denote by $\Delta^{(t)}$ the difference of the weight vector before and after the $t$-th training step.

We propose a method for explaining deep learning representations based on tracking the $\Delta^{(t)}$ values during the training stage. 
We show how it can be used to identify the most influential training examples, which can be used as a form of explanation of model decisions.
Next we propose a modified training protocol that also works in a mini-batch setting. 
Finally we propose a visual explanation that summarizes the training process of a learned representation.


\subsection{Explaining Pre-Activations by the Training History}
Deep networks compute outputs in a sequence of layers. In this work we tackle the problem of providing explanations for a the intermediate representations learned in any of these layers.
We begin our analysis for the case of single-instance stochastic training without mini-batching as in \eqref{eq:sgd}. Then we develop a deep learning method for training with specialized mini batches.

For notational simplicity we define a deep network as a composition of layers. We note that our method can also be applied to more complicated networks e.g. networks with residual connections, etc. 
For additional simplicity we limit our analysis to linear layers without bias terms, which includes fully connected layers as well as convolutional layers.
We define the \emph{output} of the $l$-th layer of a deep network 
\begin{equation}
f_l(x) = \sigma_l\left(W_l f_{l-1}(x)\right)
\end{equation}
where $\sigma_{l}$ is an activation function and $W_l,$ is a real-valued weight matrix.
We define $f_0(x):=x$.
We focus on the \emph{pre-activations} $W_l f_{l-1}(x)$ of the $l$-th layer.

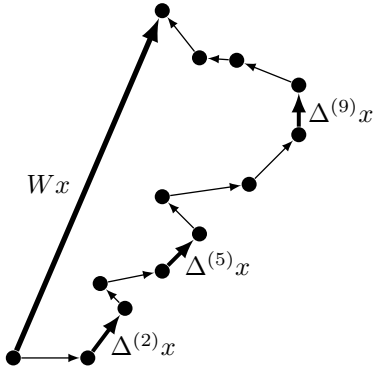
\begin{figure}[t]
\centering
\begin{tikzpicture}[every node/.style={fill, circle, inner sep=0, minimum size=0.2cm}, scale=0.66]
		\tikzset{edge/.style = {->, >=latex, line width=.5pt}}
		\node [] (0) at (0, 0) {};
		\node [] (1) at (3, 7) {};
		\node [] (2) at (1.5, 0) {};
		\node [] (3) at (2.25, 1) {};
		\node [] (4) at (1.75, 1.5) {};
		\node [] (5) at (3, 1.75) {};
		\node [] (6) at (3.75, 2.5) {};
		\node [] (7) at (3, 3.25) {};
		\node [] (8) at (4.75, 3.5) {};
		\node [] (9) at (5.75, 4.5) {};
		\node [] (10) at (5.75, 5.5) {};
		\node [] (11) at (4.5, 6) {};
		\node [] (12) at (3.75, 6.05) {};
		\draw[edge, line width=2pt](0) to node[fill=none, left, xshift=-.5em] {$Wx$} (1);
		\draw[edge] (0) to (2);
		\draw[edge, line width=1.5pt] (2) to node[fill=none, right, yshift=-.4em, xshift=.0em] {$\Delta^{(2)}x$}  (3);
		\draw[edge] (3) to (4);
		\draw[edge] (4) to (5);
		\draw[edge, line width=1.5pt] (5) to node[fill=none, right, yshift=-.4em, xshift=.0em] {$\Delta^{(5)}x$}  (6);
		\draw[edge] (6) to (7);
		\draw[edge] (7) to (8);
		\draw[edge] (8) to (9);
		\draw[edge, line width=1.5pt] (9) to node[fill=none, right, yshift=0, xshift=.2em] {$\Delta^{(9)}x$}  (10);
		\draw[edge] (10) to (11);
		\draw[edge] (11) to (12);
		\draw[edge] (12)  to (1);

\end{tikzpicture}	
\caption{Illustration of the decomposition of preactivations into individual gradient updates. The marked updates are the most substantial individual contributions to $Wx$.}
\label{fig:decomposition}
\end{figure}

In a slight abuse of notation, following \eqref{eq:core}, we define the decomposition of the weights into the contributions at every training step
\begin{equation}
W_l = \sum\limits_{t=1}^T \Delta_l^{(t)}
\label{eq:weight_decomposition}
\end{equation}
and consequently we can decompose the preactivations as
\begin{equation}
	W_lf_{l-1}(x) = \sum\limits_{t=1}^T \Delta_l^{(t)}f_{l-1}(x)
\end{equation}
as is depicted in Fig.~\ref{fig:decomposition}.
To measure the most substantial individual contributions to the preactivation, we compute the projection of the contributions onto the overall representation in the $l$-th layer
\begin{equation}
	\gamma_l^{(t)} := \left\langle W_l f_{l-1}(x), \Delta_l^{(t)} f_{l-1}(x)\right\rangle 
	\label{eq:dot}
\end{equation}
and we use large $\gamma$-values to construct explanations.
In addition to the dot product, we also consider the cosine similarity
\begin{equation}
	\bar\gamma_l^{(t)} := \gamma_l^{(t)} \cdot \left(||W_l f_{l-1}(x)||_2\cdot||\Delta_l^{(t)} f_{l-1}(x)||_2\right)^{-1}
	\label{eq:cosine}
\end{equation}
that describes the angle between the contribution and the final representation.

One special layer is the last layer, as it outputs the logits of the predictions. Hence we can contribute the predicted class probabilities to individual training steps by inspecting the decomposed preactivations there.

\subsection{Most Influential Training Examples}
We can use the $\gamma$ values to find the most influential training examples for a decision of interest.
Each $\gamma^{(t)}$ is based on exactly on $\Delta^{(t)}$ which, in turn, is based on the loss computed on exactly one training example for stochastic training without minibatching.
During training over multiple epochs, each example is used for training multiple times. In this case, we aggregate all $\gamma$ values for each training example by summing them up to get a total contribution over all training epochs.
This approach is outlined in Alg.~\ref{alg:influences}.

Note that for learning a classifier that discriminates one class vs. all the other classes, influential examples must not necessarily belong to the positive class. Perhaps even the more interesting examples belong to the other class but were misclassified during training, leading to weight updates to correct the behavior.
\begin{algorithm}[h]
\caption{Computing Influential Examples}
\label{alg:influences}
\begin{algorithmic}
\Function{compute-influence}{$f,x,\Delta_l^{(\cdot)}$}
\State $\Gamma_i = 0$ for $i = 1,\ldots, n$ \Comment{Initialize Influences}
\State Compute $f_{(l-1)}(x)$ and $W_lf_{(l-1)}(x)$
\vspace{3pt}\For{$t = 1,\ldots,T$} \Comment{Iterate Training History}
\State Identify example $x_i$ used for $\Delta^{(t)}$
\State Compute $\gamma_l^{(t)}$ with Eq. \eqref{eq:dot}
\State $\Gamma_i = \Gamma_i + \gamma_l^{(t)}$ \Comment{Sum up example importances}
\EndFor
\State \Return top-k indices in $\Gamma$
\EndFunction
\end{algorithmic}	
\end{algorithm}

%
%
\subsection{Mini-Batch-Training}
\label{sec:minibatch}
Mini-batch training can significantly decrease the duration of training by reducing the overall number of parameter updates to a model. 
Modern deep networks are almost always trained with mini-batching to allow the efficient processing of massive datasets exploiting parallel computations on GPUs.
In our case, however, it causes a serious problem: We can no longer attribute updates to individual training instances or their respective class labels.
We propose an alternative for mini-batching where we train with mini-batches of training instances that all belong to the same class. This way we can at least attribute updates to classes rather than individual examples. 
Unfortunately this causes new problems for many modern deep network architectures, as it limits the usefulness of batch normalization layers. 
The appealing property of batch normalization is that it normalizes features in a batch, such that their values lie in a  a well-defined range,
which allows efficient training without numerical instabilities, 
but also differ, which allows the model to discriminate between the classes \cite{batchnorm}.

\begin{algorithm}[tb]
\caption{Batch-Normalization with Ghost Samples}
\label{algo:ghost}
\begin{algorithmic}
\Function{ghostsample-batchnorm}{$x$, $\varepsilon = 10^{-5}$}
	\vspace{2pt}\State $\triangleright\;$ Split ghost-batch examples and compute statistics
	\State $[x_\mathrm{batch},\; x_\mathrm{ghost}] = \mathrm{split}(x)$ 
	\State $\rlap{$\mu_\mathrm{ghost},\; \mu_\mathrm{batch}$}\hphantom{[x_\mathrm{batch},\; x_\mathrm{ghost}]} = \mathrm{mean}(x_\mathrm{ghost}),\;\mathrm{mean}(x_\mathrm{batch})$ 
	\State $\rlap{$\sigma^2_\mathrm{ghost}, \;\sigma^2_\mathrm{batch}$}\hphantom{[x_\mathrm{batch},\; x_\mathrm{ghost}]} = \mathrm{var}(x_\mathrm{ghost}),\;\mathrm{var}(x_\mathrm{batch})$
	\Statex
	\State $\triangleright\;$ {Combine batch and ghost statistics}
	\State $\rlap{$\mu$} \hphantom{\sigma^2}= \mu_\mathrm{ghost} + (\mu_\mathrm{ghost} - \mu_\mathrm{batch}) / (1 + |x_\mathrm{ghost}|)$
    \State $\rlap c\hphantom{\sigma^2} = {|x_\mathrm{ghost}|} / {(1 + |x_\mathrm{ghost}|)}$
	\State $\sigma^2 = c\cdot \sigma^2_\mathrm{ghost} + \sigma^2_\mathrm{sample} $ 
	\State $\hphantom{\sigma^2=}+c\cdot (\mu_\mathrm{ghost} - \mu_\mathrm{batch})^2 / (1 + |x_\mathrm{ghost}|)]$
	\State Update running statistics with $\mu$ and $\sigma^2$
	\vspace{2pt}\State\Return $(x - \mu) / (\sigma + \varepsilon)$
\EndFunction
\end{algorithmic}
	
\end{algorithm}

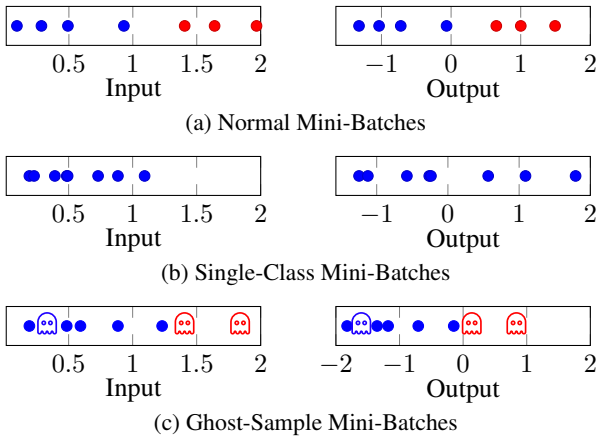
\begin{figure}
\begin{subfigure}[b]{\linewidth}%
\centering
\begin{tikzpicture}%
\begin{groupplot}[group style={group size=2 by 1,vertical sep = 30pt},height=2.1cm, width=.59\linewidth, ]
    \nextgroupplot[xlabel=Input, ymajorticks=false, xlabel style={at={(0.5,6pt)}}, yticklabels={,,},xmin=0.01, xmax=2]
    \addplot[scatter, only marks, scatter src=\thisrow{class},
      error bars/.cd, y dir=both, x dir=both, y explicit, x explicit, error bar style={color=mapped color}]
      table[x=x,y=y,x] {
	    x       xerr    y        yerr       class
	    0.28521 0.00031 0        0.000001   0
	    0.49238 0.00044 0      0.00000025 0
	    0.09346 0.00032 0 0.0000012  0
	    0.93021 0.00078 0     0.0000016  0
	    1.40567 0.00047 0 0.000002   1
	    1.63971 0.00064 0 0.0000028  1
	    1.96753 0.00063 0        0.000004   1
	};
    \nextgroupplot[xlabel=Output,ymajorticks=false, xlabel style={at={(0.5,6pt)}}, yticklabels={,,}xmin=-2, xmax=2]
    \addplot[scatter, only marks, scatter src=\thisrow{class},
      error bars/.cd, y dir=both, x dir=both, y explicit, x explicit, error bar style={color=mapped color}]
      table[x=x,y=y,x] {
	    x       xerr    y        yerr       class
	    -1.03416833 0.00031 0        0.000001   0
	    -0.72287029 0.00044 0      0.00000025 0
	    -1.32229595 0.00032 0 0.0000012  0
	    -0.06497764 0.00078 0     0.0000016  0
	    0.64945865 0.00047 0 0.000002   1
	    1.00113213 0.00064 0 0.0000028  1
	    1.49372142 0.00063 0        0.000004   1
	};
	\end{groupplot}
	\end{tikzpicture}\vspace{-6pt}
	\caption{Normal Mini-Batches}
\end{subfigure}\\[6pt]%
\begin{subfigure}[b]{\linewidth}%
\centering
\begin{tikzpicture}%
\begin{groupplot}[group style={group size=2 by 1,vertical sep = 30pt},height=2.1cm, width=.59\linewidth, ]
	\nextgroupplot[xlabel=Input, ymajorticks=false, xlabel style={at={(0.5,6pt)}}, yticklabels={,,}xmin=0, xmax=2]
    \addplot[scatter, only marks, scatter src=\thisrow{class},
      error bars/.cd, y dir=both, x dir=both, y explicit, x explicit, error bar style={color=mapped color}]
      table[x=x,y=y,x] {
	    x       xerr    y        yerr       class
	    0.88521 0.00031 0        0.000001   0
	    0.39238 0.00044 0      0.00000025 0
	    0.19346 0.00032 0   0.0000012  0
	    0.23021 0.00078 0     0.0000016  0
  	    0.48521 0.00031 0        0.000001   0
	    0.49238 0.00044 0      0.00000025 0
	    1.09346 0.00032 0 0.0000012  0
	    0.73021 0.00078 0     0.0000016  0
	};
    \nextgroupplot[xlabel=Output,ymajorticks=false, xlabel style={at={(0.5,6pt)}}, yticklabels={,,}xmin=-2, xmax=2]
    \addplot[scatter, only marks, scatter src=\thisrow{class},
      error bars/.cd, y dir=both, x dir=both, y explicit, x explicit, error bar style={color=mapped color}]
      table[x=x,y=y,x] {
	    x       xerr    y        yerr       class
	    1.08940806 0.00031 0        0.000001   0
	    -0.57591855 0.00044 0      0.00000025 0
	    -1.24809105 0.00032 0 0.0000012  0
	    -1.12390877 0.00078 0     0.0000016  0
	    -0.2622358 0.00047 0 0.000002   0
	    -0.23800759 0.00064 0 0.0000028  0
	    1.79310764 0.00063 0        0.000004   0
	    0.56564606 0.00063 0        0.000004   0

	}; 
	\end{groupplot}
	\end{tikzpicture}\vspace{-6pt}
	\caption{Single-Class Mini-Batches}
	\label{fig:ghost_samples_b}
	\end{subfigure}\\[6pt]%
\begin{subfigure}[b]{\linewidth}
\centering
 \begin{tikzpicture}
  \begin{groupplot}[group style={group size=2 by 1,vertical sep = 30pt},height=2.1cm, width=.59\linewidth, ]
	\nextgroupplot[xlabel=Input, ymajorticks=false, xlabel style={at={(0.5,6pt)}}, yticklabels={,,}xmin=0, xmax=2]
    \addplot[scatter, only marks, scatter src=\thisrow{class}, scatter/classes={
    	0={mark=*, blue}, 1={mark=ghost_red},2={mark=ghost_blue}},
      error bars/.cd, y dir=both, x dir=both, y explicit, x explicit, error bar style={color=mapped color}]
      table[x=x,y=y,x] {
	    x       xerr    y        yerr       class
	    0.88521 0.00031 0        0.000001   0
	    0.59238 0.00044 0      0.00000025 0
	    0.19346 0.00032 0   0.0000012  0
	    1.23021 0.00078 0     0.0000016  0

  	    0.48521 0.00031 0        0.000001   0
	    0.3321 0.00078 0     0.0000016  2	   
	    1.40567 0.00047 0 0.000002   1
	    1.83971 0.00064 0 0.0000028  1
	};
    \nextgroupplot[xlabel=Output,ymajorticks=false, xlabel style={at={(0.5,6pt)}}, yticklabels={,,},xmin=-2, xmax=2]
    \addplot[scatter, only marks, scatter src=\thisrow{class}, scatter/classes={
    	0={mark=*, blue}, 1={mark=ghost_red},2={mark=ghost_blue}},
      error bars/.cd, y dir=both, x dir=both, y explicit, x explicit, error bar style={color=mapped color}]
      table[x=x,y=y,x] {
	    x       xerr    y        yerr       class
	    -0.70530926 0.00031 0        0.000001   0
	    -1.17892468 0.00044 0      0.00000025 0
	    -1.82412723 0.00032 0   0.0000012  0
	    -0.14731548 0.00078 0     0.0000016  0
  	    -1.35225857 0.00031 0        0.000001   0
	    -1.5998946 0.00078 0     0.0000016  2	   
	     0.13646884 0.00047 0 0.000002   1
	    0.83847353 0.00064 0 0.0000028  1
	};  
  \end{groupplot}
\end{tikzpicture}\vspace{-6pt}
	\caption{Ghost-Sample Mini-Batches}	
	\label{fig:ghost_samples_c}
\end{subfigure}
\caption{Illustration of the different batch-normalization variants, color indicates class labels. (b) When we construct batches with a single class, the outputs span the whole range. An imaginary next batch of the other class would to the same, hindering training. (c) Ghost samples solve this problem, as mean and variance are also computed on the ghost batch.}
\label{fig:ghost_samples}
\end{figure}

But when we train with batches of a single class, the batch normalization layer forces feature values to differ that should be similar, as illustrated in the Fig.~\ref{fig:ghost_samples_b}. We introduce \emph{ghost samples} to the mini batch to prevent this problem: These ghost samples are training instances sampled from all classes according to the class prior probabilities. We compute the batch normalization statistics on the ghost samples and the normal batch separately.
Then we combine the statistics by downweighting  the examples in the mini batch to only have a total weight corresponding to one example in the ghostbatch as outlined in Algorithm~\ref{algo:ghost}. This way all examples contribute to the mean and variance estimates, but the same-class batch has only little influence, as illustrated in the third row of Fig.~\ref{fig:ghost_samples}.

When evaluating the loss of our model, however, we only use the samples in the original minibatch without the ghost samples. This way normalization is based on all classes, while loss is based only on a single class.
Now the model's gradients and consequently the parameter updates are mostly based on the single-class batch.
This allows us to still attribute individual training steps to a set of examples of a single class. Instead of presenting the examples that were most influential, we can now present the most influential minibatches.
When computing the overall contribution of an individual example, the contributions of a minibatch are attributed to all its training examples equally. This way we can still aggregate the contributions by summation.
While we could also aggregate contributions in a normal mini-batch, in practice this yields unusable values. With one-class minibatches, however, the results are better. We believe this is due to the lower variance in the batch.
In the next section we present an alternative visual explanation that aggregates class-wise rather than example-wise.

\subsection{Visualizing Most Influential Classes}
Privacy Requirements may prevent us from explaining decisions to users by showing the labeled training data as evidence, e.g. in situations where training data contains personal information like for automated credit scoring or in medical applications. In many situations we can still provide explanations based on aggregated information, e.g. when the number of training examples is sufficiently large that aggregation does not leak personal information. 
Displaying aggregated information has the additional advantage that it shows a more complete picture of the training process, whereas showing individual training instances as explanations shows only a very small part of the process and the interpretation of the instances is very subjective.

For visualizing aggregated statistics, we  propose to show ridge plots to aggregate all $\gamma_l^{(t)}$ values grouped by layer and training labels. These plots for a single layer are structured as follows: We use one subplot for each class of interest. Its $x$-axis shows the cosine similarities $\bar \gamma_l^{(t)}$ between the individual contributions and the total preactivations according to \eqref{eq:cosine}, on the $y$-axis we show a density estimate of how frequently this cosine similarity appeared during training. Instead of showing raw counts, we weight each of them by the magnitude of the update $||\Delta_l^{(t)} f_{l-1}(x)||_2$. We decided against the use of histograms and used kernel density estimates instead, mostly for aesthetic reasons.
By controlling the kernel bandwidth parameter we can control the amount of smoothing to the plotted density. Throughout our analysis we use a bandwidth of $0.1$ which gives a good tradeoff between accuracy of the density estimates and readability. This approach is outlined in Alg.~\ref{alg:ridge}. For visual guidance we highlight the predicted class and only show the classes with the top-10 predicted probabilities. You can find an example ridge plot in Figure~\ref{fig:example}: A high influence of a certain class is indicated by large amount of probability mass at high similarities, as is the case in the example for the target class "frog".
\begin{algorithm}[h]
\caption{Computing Ridge Plots}
\label{alg:ridge}
\begin{algorithmic}
\Function{visualize-influence}{$f,x,\Delta_l^{(\cdot)}$}
\vspace{2pt}\For{\textbf{each} class $y$} \Comment{Initialize statistics}
\State $\Gamma_y = \emptyset$ 
\EndFor
\State Compute $f_{(l-1)}(x)$ and $W_lf_{(l-1)}(x)$
\vspace{3pt}\For{$t = 1,\ldots,T$} \Comment{Iterate training history}
\State Identify class $y_i$ used for $\Delta^{(t)}$
\State Compute $\bar\gamma_l^{(t)}$ with Eq. \eqref{eq:cosine}
\vspace{2pt}\State $\triangleright\;$Store cos-similarity and weight for class $y_i$
\State $\Gamma_{y_i} = \Gamma_{y_i} \cup \{(\bar\gamma_l^{(t)},||\Delta_l^{(t)} f_{l-1}(x)||_2)\}$
\EndFor
\State plots $= \emptyset$
\For{\textbf{each} class $y$} 
\State $\triangleright\;$Plot cosine similarities and weights from $\Gamma_y$
\State plots = plots $\:\cup\: \{\mathrm{kdePlot}(\Gamma_y, \textrm{kernel-bw}=0.1)\}$ 
\EndFor
\State \Return plots
\EndFunction
\end{algorithmic}	
\end{algorithm}
Note that inspecting the output layer in these plots goes beyond obtaining a class probability at prediction time: They allow to understand how these class probabilities came about.

\section{Experiments}
\label{sec:experiments}
In this section we evaluate our method for two different classification tasks and analyze the computational overhead. We evaluate our custom minibatch training and how it allows us to provide explanations for individual decisions.
For every figure presented in this section you can find an interactive version as well as additional examples in the supplementary materials. The source code can be found at \url{https://github.com/Whadup/xai_tracking}.

\subsection{Explanations for Image Classification}
We investigate our method for a popular image classification benchmark, the Cifar-10 dataset. 
We train a convolutional neural network that uses the VGG-16 architecture -- 13 convolution layers followed by 3 fully connected layers -- and train it with our mini-batch training algorithm with a batch size of 128 examples, of which 40 examples are the ghost sample containing 4 images of each class. Note that the classes are balanced in the train- as well as test-data, so the ghost sample also should also have this property. We train using vanilla stochastic gradient descent for 35 epochs with an initial learning rate of $0.01$ which is decreased by a factor of $0.1$ every 10 epochs. 
Our model achieves a test accuracy of $82\%$, which is not state-of-the-art, but sufficiently high to demonstrate the usefulness of our explanation approach.

\begin{figure}
\includegraphics[keepaspectratio, width=\linewidth]{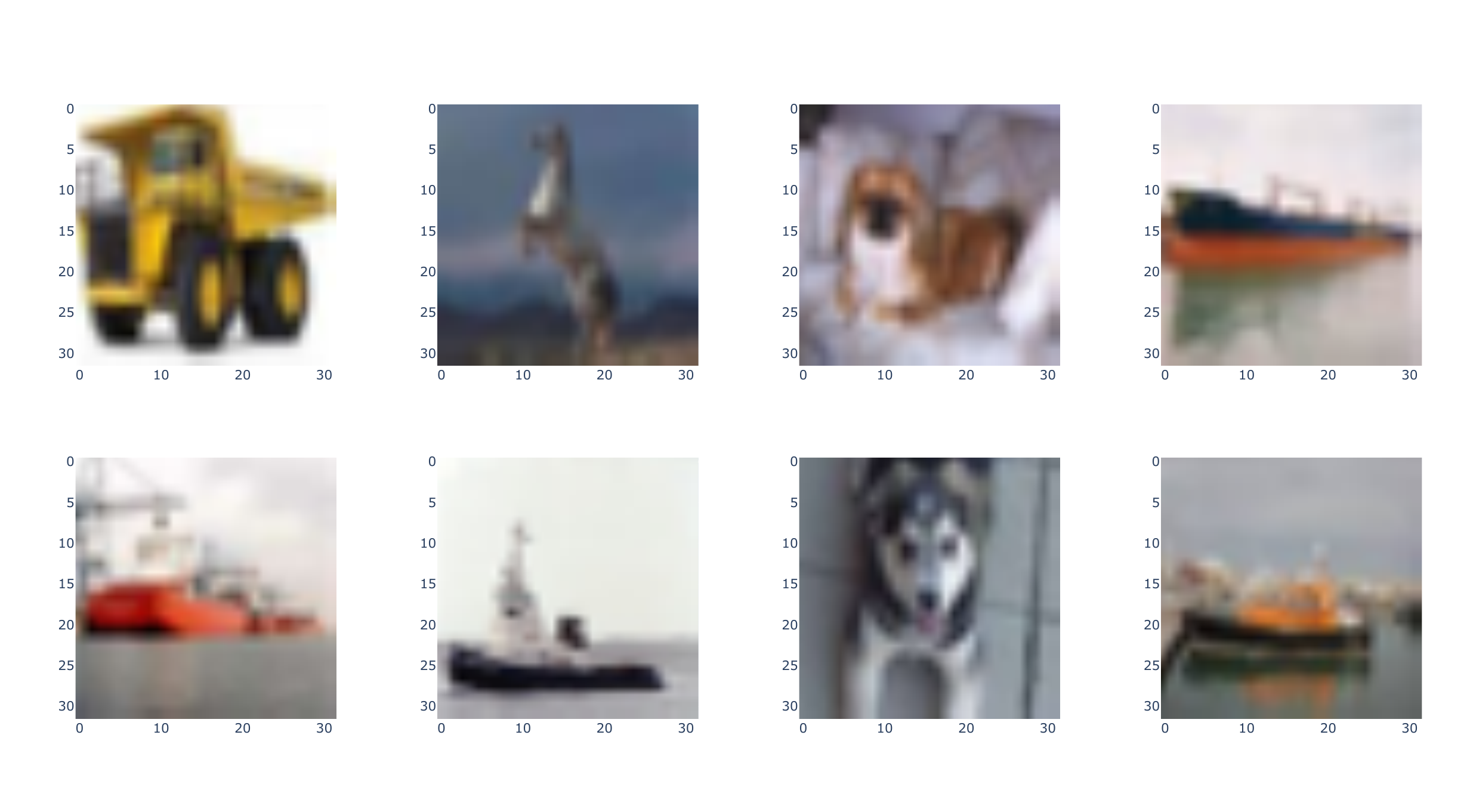}
\caption{Highly influential images for the 14th layer from a variety of classes leave the practitioner confused. Input image in top left position for reference.}
\label{fig:chaos}
\end{figure}

\begin{figure*}
\begin{subfigure}[b]{0.32\textwidth}
\includegraphics[keepaspectratio, width=\linewidth]{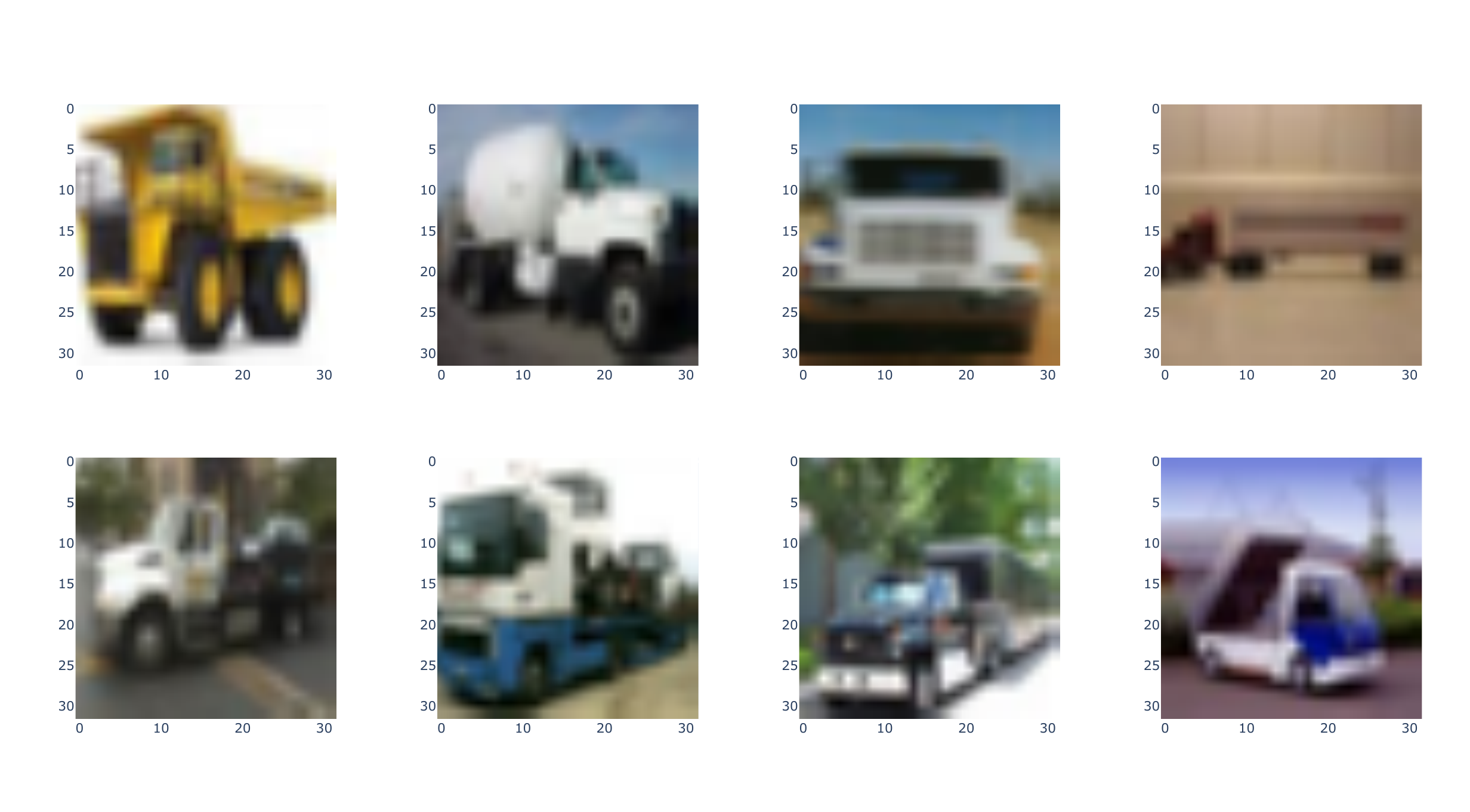}
\caption{13th Layer}
\label{fig:image_truck_a}
\end{subfigure}
\begin{subfigure}[b]{0.32\textwidth}
\includegraphics[keepaspectratio, width=\linewidth]{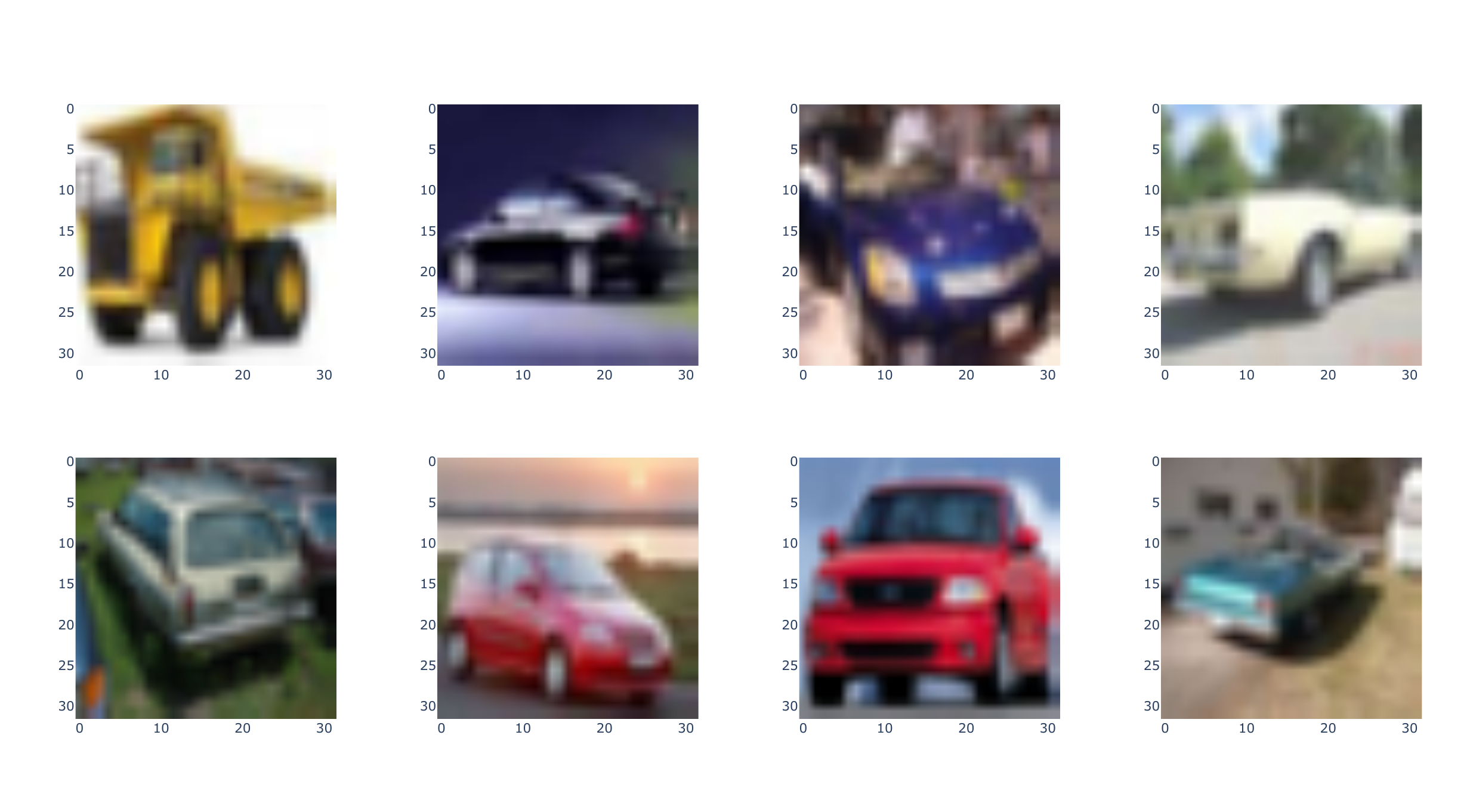}
\caption{10th Layer}
\label{fig:image_truck_b}
\end{subfigure}
\begin{subfigure}[b]{0.32\textwidth}
\includegraphics[keepaspectratio, width=\linewidth]{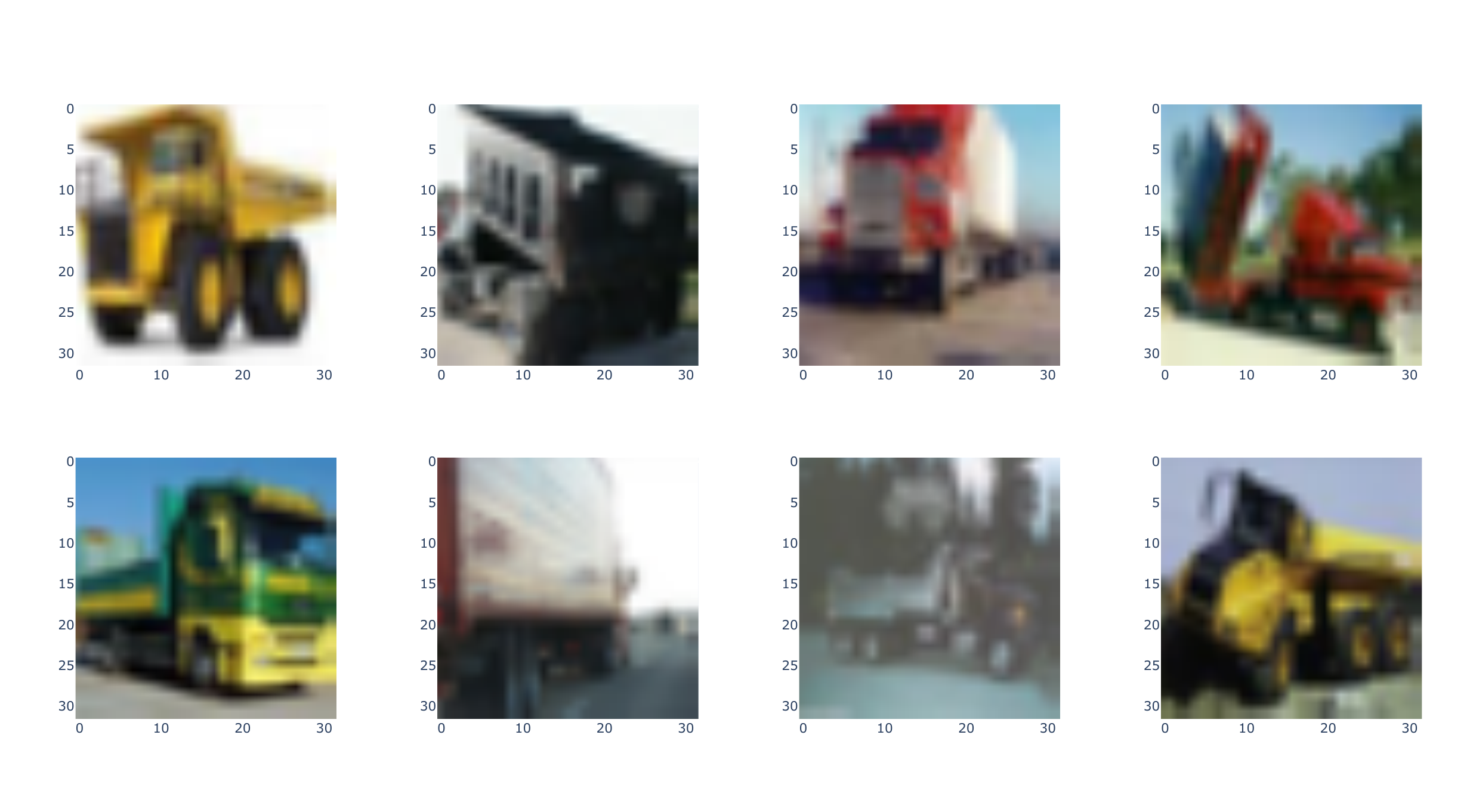}
\caption{4th Layer}
\label{fig:image_truck_c}
\label{fig:image_truck}
\end{subfigure}

\caption{The 7 most influential training examples for the 3 layers of the VGG-16 network. 
Top left image is the input image for reference.
We see a high influence of the class "car" on the representation in the middle layers, here e.g. in the 10th layer.
The earlier convolutional layers, eg. the 4th layer, is influenced mostly by very similar images, i.e. trucks photographed from an angle, whereas the linear layer, here the 13th layer, is influenced by images of trucks shot from a variety of angles. 
}
\label{fig:ridge_image}
\end{figure*}

\begin{figure*}
\begin{subfigure}[b]{0.32\textwidth}
\includegraphics[keepaspectratio, width=\linewidth]{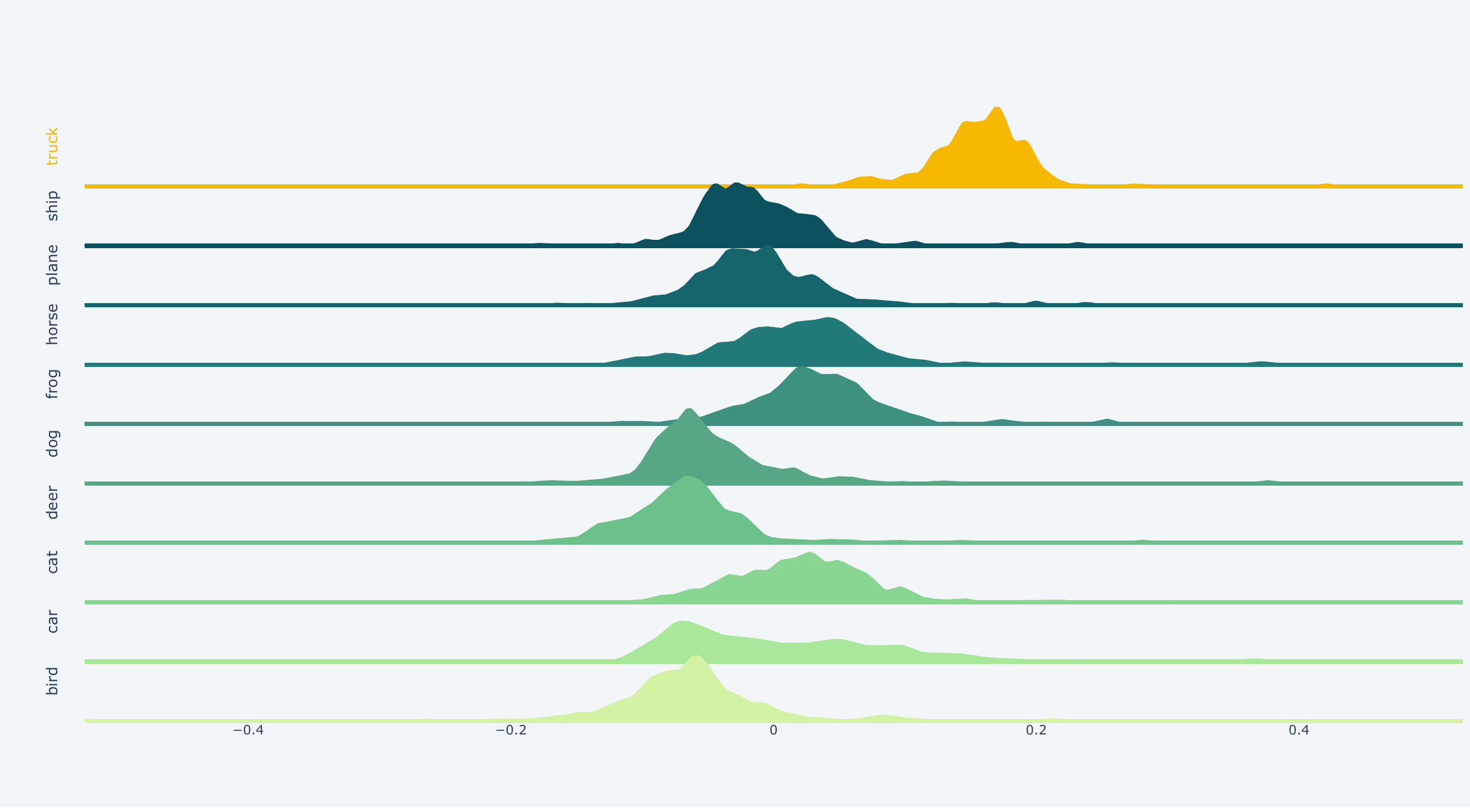}
\caption{13th Layer}
\label{fig:ridge_truck_a}
\end{subfigure}
\begin{subfigure}[b]{0.32\textwidth}
\includegraphics[keepaspectratio, width=\linewidth]{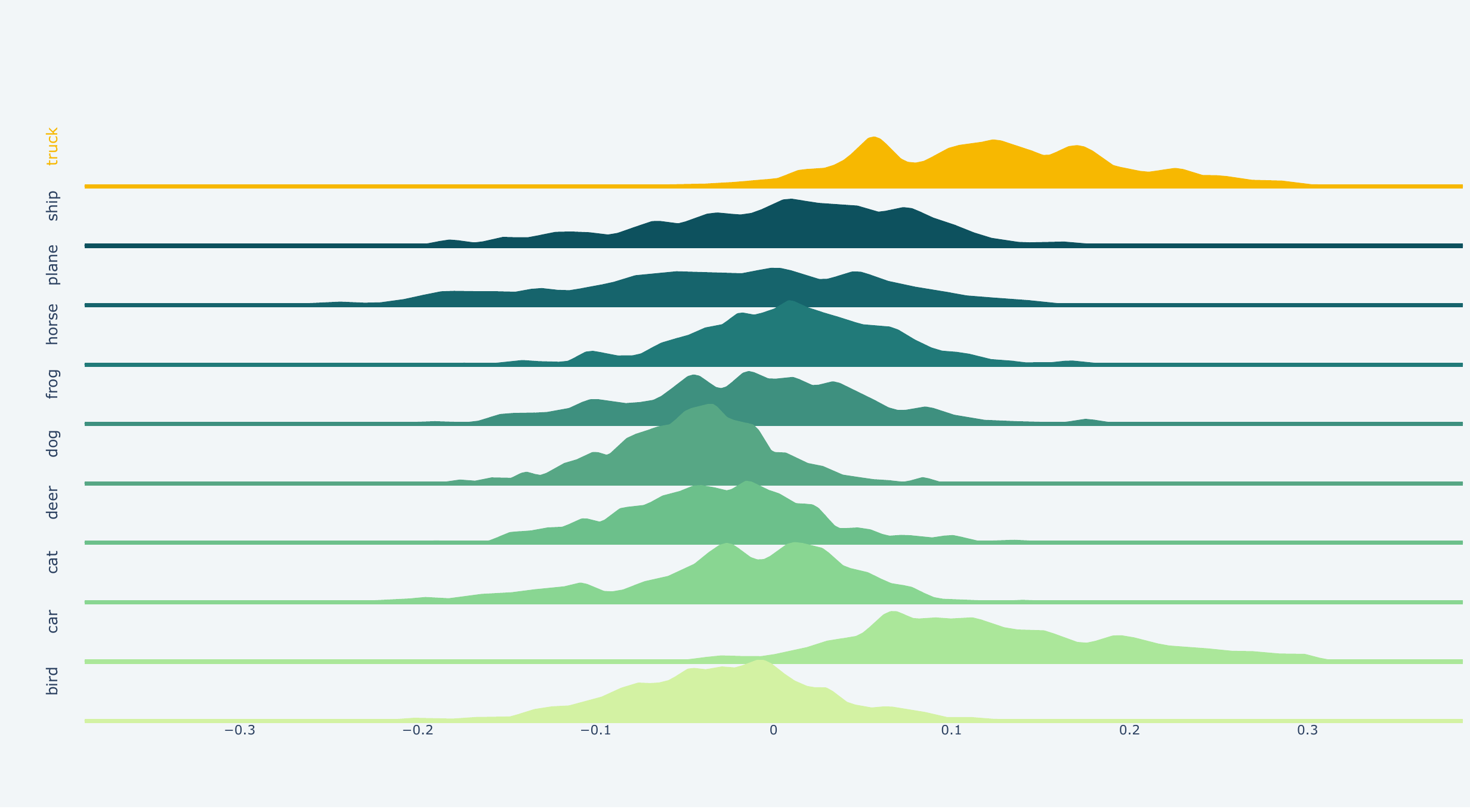}
\caption{10th Layer}
\label{fig:ridge_truck_b}
\end{subfigure}
\begin{subfigure}[b]{0.32\textwidth}
\includegraphics[keepaspectratio, width=\linewidth]{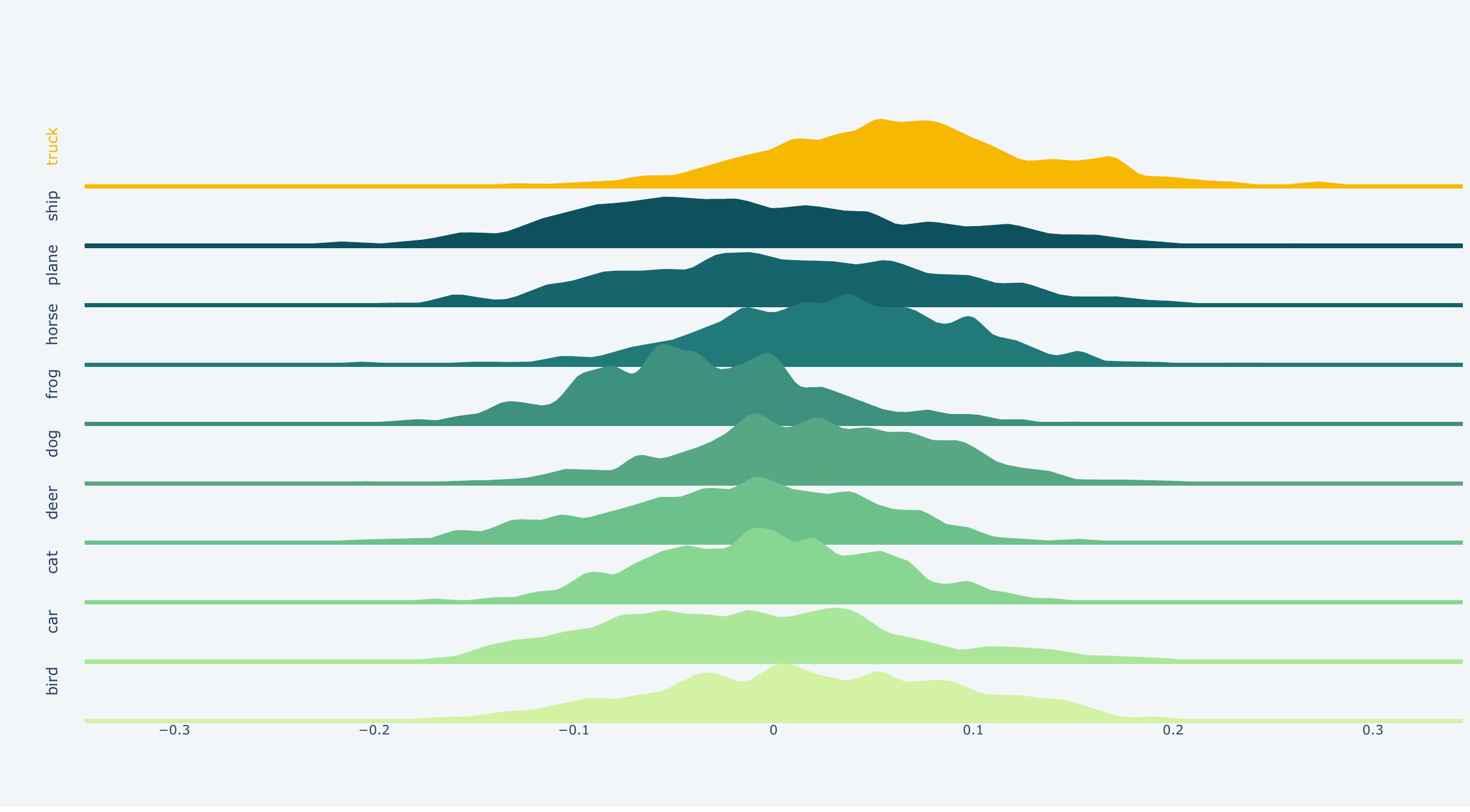}
\caption{4th Layer}
\label{fig:ridge_truck_c}
\end{subfigure}

\caption{Ridge Plot for the 3 layers of the VGG-16 network. 
We see a high influence of the class "car" on the representation in the middle layers, here e.g. in the 10th layer.
The earlier layers are not as descriptive of individual classes, eg. the 4th layer.
}
\label{fig:ridge_truck}
\end{figure*}

We investigate both types of visual explanations proposed in this work, the most influential training examples, see e.g. Fig.~\ref{fig:image_truck} and ridge plots of all contributions, e.g. in Fig.~\ref{fig:ridge_truck}.
Both forms of visualization show changing behavior along the depth of the network:
In the ridge plots, we see that later layers are more specific to classes and hence seem more important for explanations, whereas earlier layers do not differ substantially. For instance in the 13th linear layer, Fig.~\ref{fig:image_truck_a}, we see a clear peak of high similarities i.e. high influence for the true class "truck", whereas in the earlier 4th convolution layer, Fig.~\ref{fig:image_truck_c}, the classes behave more homogeneous.
When looking at the most-influential training images, the earlier layers show only images of the predicted class, as in Fig.~\ref{fig:image_truck_c}. But later layers sometimes show very mixed images, as e.g. the 14th layer in Fig~\ref{fig:chaos}. Further analysis reveals that these outputs are triggered by weight updates very early in the training process when the model was still untrained and the optimization performed large weight updates to learn. Future work should investigate this behavior further and try to find remedies. Other layers, like the 13th layer in Fig~\ref{fig:image_truck_a}, again only show images of the target class, but we do note a higher variety in the images , for instance in the perspective of the images, in comparison to the earlier layers.
All in all, this changing behavior along the depth of the network is in line with previous results, that illustrate how convolution filters match more complex patterns corresponding to the classes in later layers, whereas the earlier layers match basic concepts like edges and textures \cite{Erhan/etal/2009a}.

Furthermore we can see that related classes influence each others representation, e.g. truck and car or plane and bird are often both influential on middle layers. This phenomenon is present in the ridge plots (see eg. Fig.~\ref{fig:ridge_truck_b}) as well as the most influential examples (see e.g. Fig.~\ref{fig:image_truck_b}). Additional examples can be found in the supplementary materials.


\subsection{Explanations for Graph Classification}
In this section, we investigate explanations for a graph neural network. We work with the ogbg-ppa dataset that contains undirected protein association neighborhoods from 37 different taxonomic groups which is the prediction target \cite{Hu/etal/2020,Szklarczyk/etal/2019,Zitnik/etal/2019}. Each node represents a protein and each edge represents an interaction and is described by 7 features.
We use the official train-test split, where the training data contains x graphs and the test data contains y graphs.
Unlike in the previous section, the classes are not perfectly balanced. We use a graph convolutional neural network based on the model proposed by Xu et al. \shortcite{Xu/etal/2019} that uses 5 layers of graph convolutions and a hidden dimensionality of 300. The edge features are transformed to 300-dimensional incoming node features \cite{Hu/etal/2020}.

The graph convolution transforms the features of a node $v$ by
\begin{equation}
	f_l(x_v) = \mathrm{MLP}_l\left(\sum_{(u,v) \in \mathcal E} \max[x_u + W_l \phi(u,v), 0]\right)
\end{equation}
where $\phi(u,v)$ denotes the edge features of the edge from $u$ to $v$.
We use the edge transformation $W_l \phi(u,v)$ as well as all linear layer in $\mathrm{MLP}_l$ to compute explanations.

We use SGD training with Nesterov momentum of 0.9 and an initial learning rate of 0.001 that is decreased by factor 0.1 after 10 and 40 epochs.
Our model trained with single-class minibatches of size 32 achieves an accuracy of 71.5\% on the test data, which places is comparable to performances of the baselines on the current public leaderbord for the ogbg-ppa dataset.

\begin{figure*}
\begin{subfigure}[b]{0.32\textwidth}
\includegraphics[keepaspectratio, width=\linewidth]{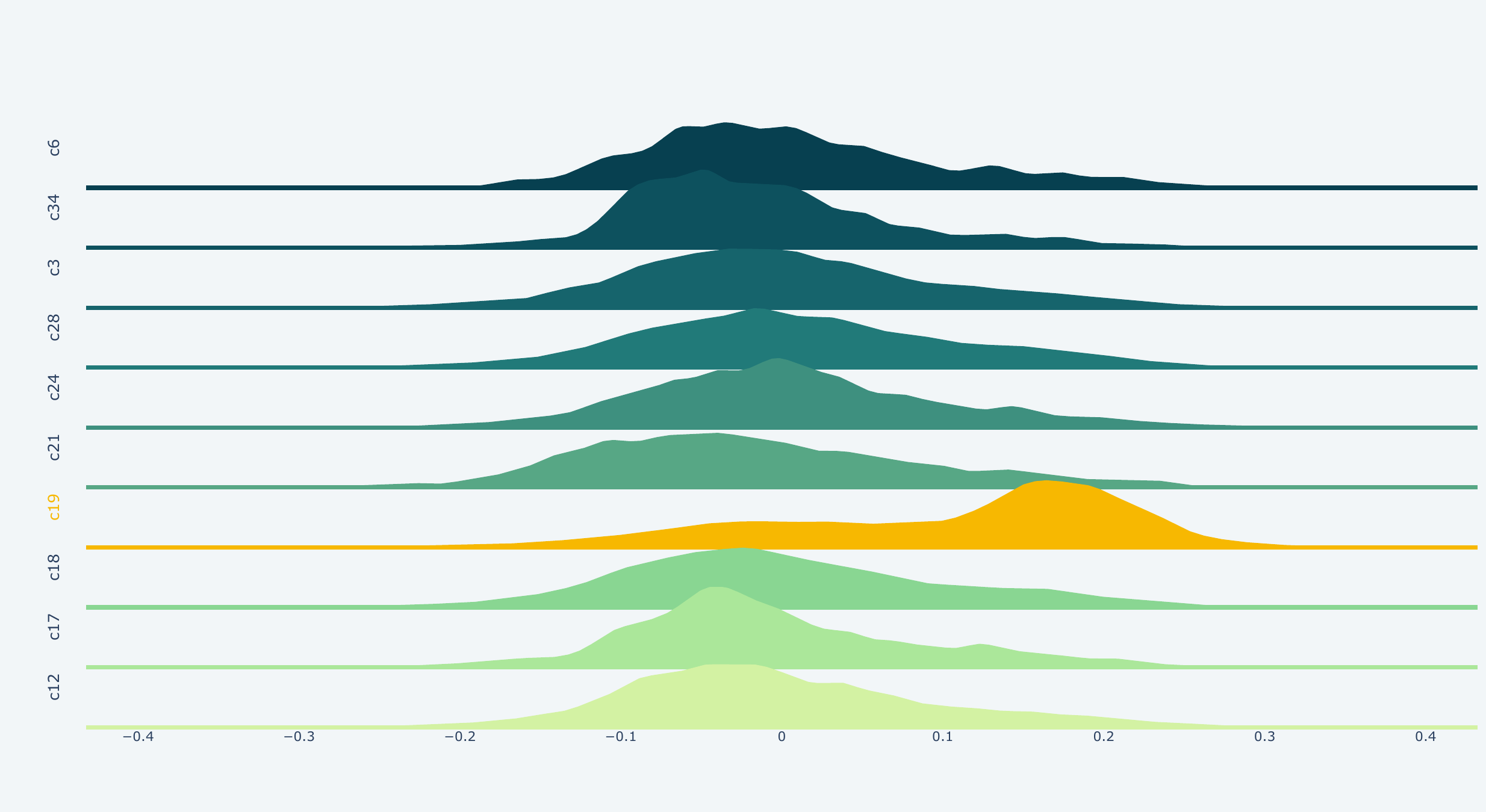}
\caption{Logit Layer}
\label{fig:ridge_graph_a}
\end{subfigure}
\begin{subfigure}[b]{0.32\textwidth}
\includegraphics[keepaspectratio, width=\linewidth]{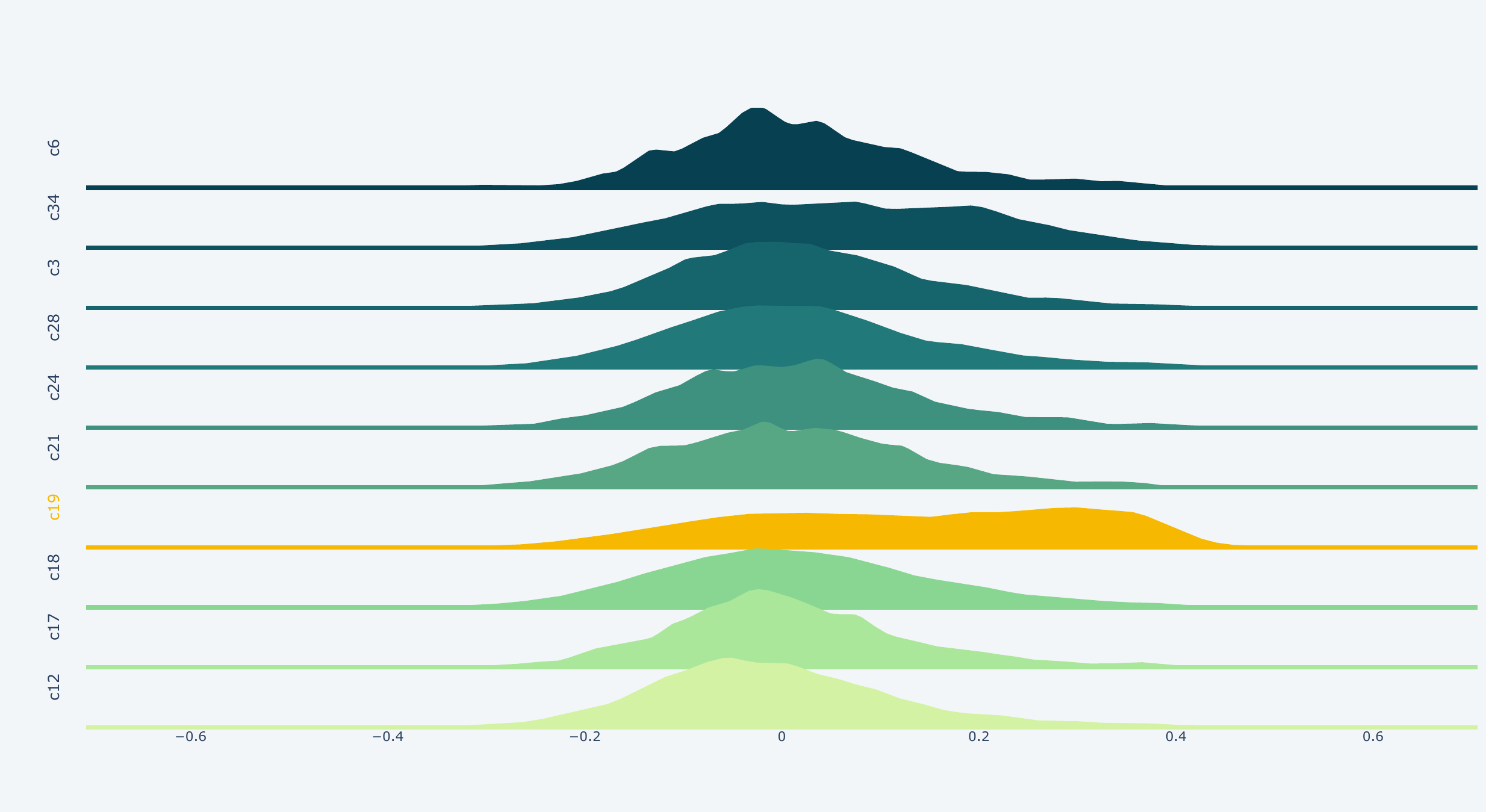}
\caption{14th Layer}
\label{fig:ridge_graph_b}
\end{subfigure}
\begin{subfigure}[b]{0.32\textwidth}
\includegraphics[keepaspectratio, width=\linewidth]{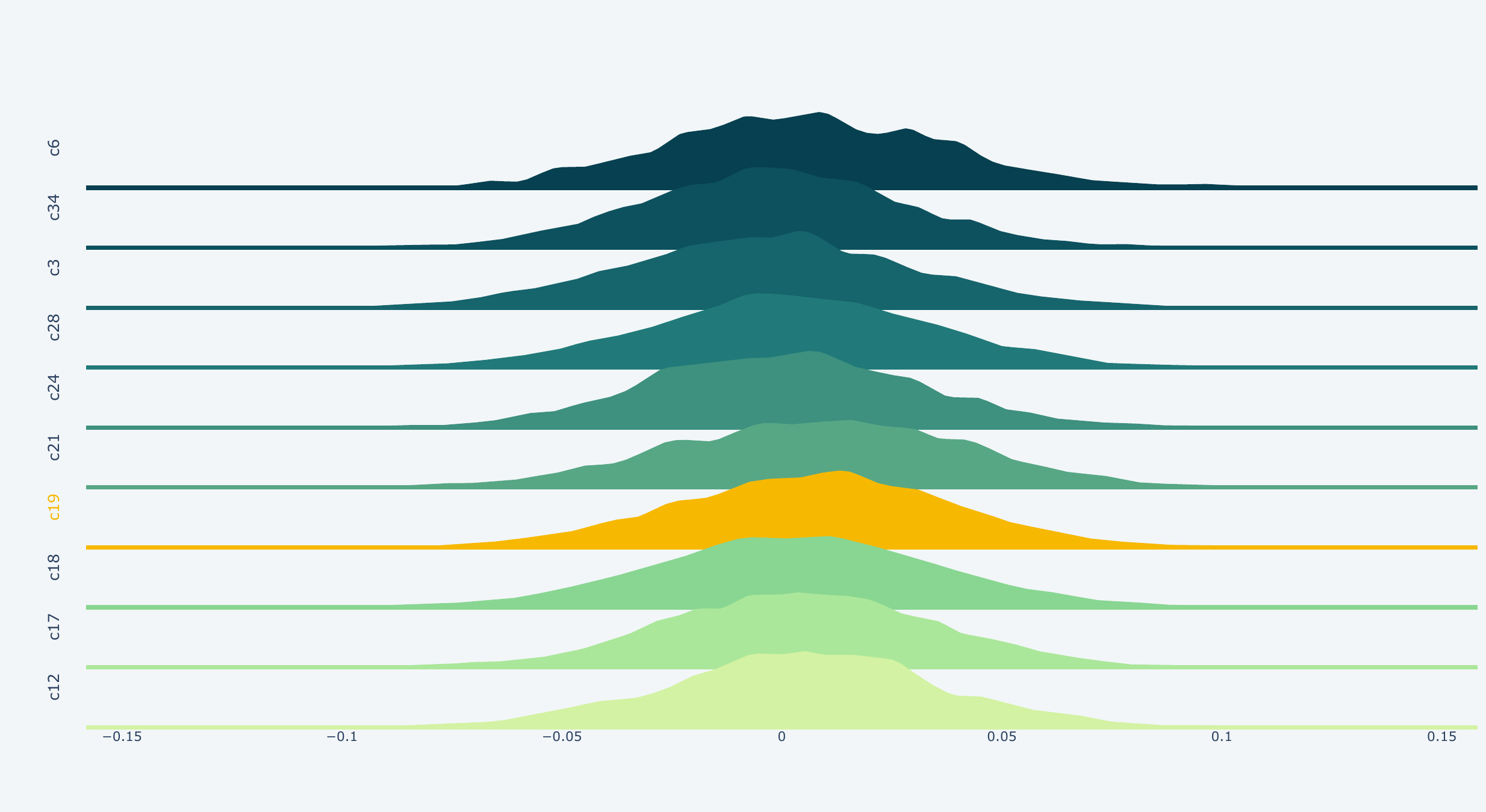}
\caption{12th Layer}
\label{fig:ridge_graph_c}
\end{subfigure}

\caption{Ridge Plot for the output layer, the 14th layer and the 12th layer. We see that while the representation at the 14th layer can be attributed to examples of the correct class, the representations in the 12th layer and in all prior layers look unspecific to a class. This indicates a poor choice of network architecture that does not utilize all layers for classification.}
\label{fig:ridge_graph}
\end{figure*}

We only inspect the ridge plots in this application, as images of graphs with more than 100 nodes and 6 numerical edge features are very hard to interpret, particularly without domain knowledge.
The visualizations for this graph neural network reveal, that all but the last two layers have almost uniform behavior throughout the classes, as e.g. shown in Fig.~\ref{fig:ridge_graph_c}. This indicates that the network is not benefitting from its depth, as the intermediate representations are not discriminative for the target class. To verify this hypothesis, we train a shallow model that uses only a single graph convolutional layer fed into a linear output layer. Indeed this experiment reveals that the shallow model also achieves 71\% test accuracy. This highlights another possible use-case of our method for machine learning engineers who can use the approach for debugging machine learning models.
%
%


%

\subsection{Computational Overhead during Training}
Biggest roadblock for implementing this system seems to be the additional overhead during training. In this section, we report the impact on runtime and memory or storage. All reported runtimes are measured on a NVIDIA DGX A100 machine using a single A100 GPU with 40GB memory.

\begin{table}
\caption{Computational Overhead Statistics for the Previous Experiments}
\label{table:overhead}
\centering
\begin{tabular}{lcc}
\toprule
Metric & cifar10 & ogp-ppa \\ \midrule
Model Size & 129MB & 7.1MB\\
Number of Weight Updates & 11,685 & 278,500\\
Runtime per Epoche & 72sec & 264s\\
Relative Overhead Runtime & +453\% & +54\%\\
Disk Storage Demand & 1.5TB & 1.9TB \\
Explanation Time & 10min &  66min \\
\bottomrule
\end{tabular}
	
\end{table}

As we can see in Table~\ref{table:overhead}, tracking all weight updates requires 1.5 and 1.9 terabytes of disk space. While this seems like a lot, it also is small enough to fit on modern compute architecture.
But, generating explanations is expensive, as we have to load these terabytes of parameters efficiently from disk onto GPU for one-time use. These I/O operations are the current computational bottleneck, explaining the long running times of generating explanations. Note that the time per explanation decreases when we generate a batch of explanations, as we only need to load the weight changes once per batch rather than once per example.

At training time, the overhead of storing all weight updates to disk on the training time depends on the model size: For the large VGG-16 model the impact is very high, mostly due to the large linear layers with over 4mio parameters each, whereas for the small graph neural network it is not severe.
At the moment we naively store all weights updates uncompressed to disk, which is probably very wasteful and makes explanation rather slow, because massive amounts of weights have to be transferred from disk to GPU. 
Different options for making this more efficient come to mind: We can for instance approximate the weight updates of large linear layers with low-rank approximations. For standard SGD-training, these updates are indeed low-rank as the weight update is computed as an outer product of input and gradient and thus its rank is at most the batch-size. An alternative to low-rank approximations are random projections as e.g. Garima et al. \shortcite{Garima/etal/2020a} propose.


\section{Conclusion and Future Research}

We have proposed an addition to the pool of explanation methods. While many other methods explain, how the model computed a decision, our method asks why the model learned to classify this way. Hence it is meant to complement existing methods to provide evidence from the training set that decisions are sound.
Our approach is extremely \emph{simple}: We track all weight updates during training which allows us to decompose the hidden activations of neural networks into a sum over training steps. This way we can attribute the representation to individual training examples. This method is \emph{general}, it can be wrapped around any optimization algorithm and covers a wide range of deep models. We have proposed two types of \emph{visual explanations}: One based on most-influential individual training instance, the other based on aggregated statistics over all training steps.

Our explanations do not attempt to ``open the black box", instead they find evidence for the decision in the training process. We can provide insights which training examples actually shaped the intermediate representations for a given example the most (Alg.~\ref{alg:influences}), and which classes influenced the intermediate representations most (Alg.~\ref{alg:ridge}).

We can answer the question ``Why did the model learn to predict $y$?'' with an explanation like in Fig.~\ref{fig:example}, optionally -- depending on the target audience -- equipped with a brief description of neural networks and their weight-decomposition \eqref{eq:weight_decomposition}. Unlike other approaches, the explanations are directly tied to the training \emph{process}, not only to the final model \cite{Koh/Liang/2017}, or the training data \cite{Yeh/etal/2018a} or even nothing but the model architecture \cite{Adebayo/etal/2018a}.

Additionally, we have seen a possible application of this approach for machine learning engineers debugging a model when our method uncovered that a deep graph convolutional neural network did not actually exploit its depth when computing predictions.

Currently our approach only analyzes the linear maps in a feed forward network. While this includes many popular machine learning models like CNNs or graph neural networks, it does exclude two popular architecture choices: gated recurrent networks and transformer architectures with their self-attention layers. How to extend our approach to these models remains an open question.



\bibliography{aaai.bib}

\end{document}